\documentclass[11pt,a4paper,dvipsnames]{article}
\usepackage[hyperref]{acl2019}
\usepackage{times}
\usepackage{latexsym}
\usepackage{booktabs}
\usepackage{amsmath}
\usepackage{graphicx}
\usepackage{url}
\usepackage{marvosym}
\usepackage{xcolor}
\definecolor{cadmiumgreen}{rgb}{0.0, 0.42, 0.24}
\definecolor{cornellred}{rgb}{0.7, 0.11, 0.11}

\aclfinalcopy 

\DeclareMathOperator*{\argmax}{arg\,max}

\DeclareMathOperator*{\softmax}{softmax}
\DeclareMathOperator*{\sigmoid}{sigmoid}
\DeclareMathOperator*{\affine}{affine}
\newcommand{\bs}[1]{\boldsymbol{#1}}

\title{Joint Learning of Named Entity Recognition and Entity Linking}

\author{Pedro Henrique Martins\textsuperscript{\Neptune} \quad
        Zita Marinho\textsuperscript{\Moon\Scorpio} \and
        Andr\'e F.T. Martins\textsuperscript{\Neptune\Saturn} \\
\textsuperscript{\Neptune}Instituto de Telecomunica\c{c}\~oes~
\textsuperscript{\Moon}Priberam Labs~
\textsuperscript{\Scorpio}Institute of Systems and Robotics~ 
\textsuperscript{\Saturn}Unbabel\\
\href{mailto:pedrohenriqueamartins@gmail.com}{pedrohenriqueamartins@gmail.com},\quad
\href{mailto:zita.marinho@priberam.pt}{zita.marinho@priberam.pt}, \\
\href{mailto:andre.martins@unbabel.com}{ andre.martins@unbabel.com}.}

\date{}

\begin{document}
\maketitle
\begin{abstract}
Named entity recognition (NER) and entity linking (EL) are two fundamentally related tasks, since in order to perform EL, first the mentions to entities have to be detected. However, most entity linking approaches disregard the mention detection part, assuming that the correct mentions have been previously detected. In this paper, we perform joint learning of NER and EL to leverage their relatedness and obtain a more robust and generalisable system. For that, we introduce a model inspired by the Stack-LSTM approach \cite{Dyer_stack_2015}.
We observe that, in fact, doing multi-task learning of NER and EL improves the performance in both tasks when comparing with models trained with individual objectives. Furthermore, we achieve results competitive with the state-of-the-art in both NER and EL.
\end{abstract}

\section{Introduction}

In order to build high quality systems for complex natural language processing (NLP) tasks, it is useful to leverage the output information of lower level tasks, such as named entity recognition (NER) and entity linking (EL). Therefore NER and EL are two fundamental NLP tasks.

NER corresponds to the process of detecting mentions of named entities in a text and classifying them with predefined types such as person, location and organisation. However, the majority of the detected mentions can refer to different entities as in the example of Table~\ref{example}, in which the mention ``Leeds'' can refer to ``Leeds'', the city, and ``Leeds United A.F.C.'', the football club. To solve this ambiguity EL is performed. It consists in determining to which entity a particular mention refers to, by assigning a knowledge base entity id. 

\renewcommand{\arraystretch}{1.4}{
\begin{table}[ht!]
\begin{centering}\small
\begin{tabular}{lll}
\hline 
\multicolumn{3}{c}{Leeds' Bowyer fined for part in fast-food fracas.} \\
\hline
& NER & EL \\
\hline
Separate &  Leeds-ORG &  \textcolor{cornellred}{\textsf{Leeds}}\\
Joint  &  Leeds-ORG & \textcolor{cadmiumgreen}{\textsf{Leeds\_United\_A.F.C.}}\\
\hline 
\end{tabular}
\par\end{centering}
\centering{}\caption{Example showing benefits of doing joint learning. Wrong entity in red and correct in green.}
\label{example}
\end{table}} 

In this example, the knowledge base id of the entity ``Leeds United A.F.C.'' should be selected.

In real world applications, EL systems have to perform two tasks: mention detection or NER and entity disambiguation. However, most approaches have only focused on the latter, being the mentions that have to be disambiguated given.

In this work we do joint learning of NER and EL in order to leverage the information of both tasks at every decision. Furthermore, by having a flow of information between the computation of the representations used for NER and EL we are able to improve the model. 

One example of the advantage of doing joint learning is showed in Table~\ref{example}, in which the joint model is able to predict the correct entity, by knowing that the type predicted by NER is Organisation.

This paper introduces two main contributions:
\begin{itemize}
    \item A system that jointly performs NER and EL, with competitive results in both tasks.
    \item A empirical qualitative analysis of the advantage of doing joint learning vs using separate models and of the influence of the different components to the result obtained.
\end{itemize}

\renewcommand{\arraystretch}{1.27}{
\begin{table*}[ht]
\begin{centering}\small
\resizebox{\linewidth}{!}{%
\begin{tabular}{llll|l}
\hline 
Action & Buffer & Stack & Output & Entity \tabularnewline
\hline 
 & [Obama, met, Donald, Trump] & [] & [] &  \tabularnewline
\texttt{Shift} & [met, Donald, Trump] & [Obama] & [] & \tabularnewline
\texttt{Reduce-PER} & [met, Donald, Trump] & [] & [(Obama)-PER] & \textsf{Barack\_Obama}\tabularnewline
\texttt{Out} & [Donald, Trump] & [] & [(Obama)-PER, met] & \textsf{Barack\_Obama}\tabularnewline
\texttt{Shift} & [Trump] & [Donald] & [(Obama)-PER, met] & \textsf{Barack\_Obama}\tabularnewline
\texttt{Shift} & [] & [Donald, Trump] & [(Obama)-PER, met] & \textsf{Barack\_Obama}\tabularnewline 
\texttt{Reduce-PER} & [] & [] & [(Obama)-PER, met,  & \textsf{Barack\_Obama}, \tabularnewline
& & & (Donald Trump)-PER]  &  \textsf{Donald\_Trump} \tabularnewline
\hline 
\end{tabular}}
\par\end{centering}
\centering{}\caption{Actions and stack states when processing sentence ``Obama met Donald Trump''. The predicted types and detected mentions are contained in the Output and the entities the mentions refer to in the Entity.}
\label{example_stack}
\end{table*}
}

\section{Related work}

The majority of NER systems treat the task has sequence labelling and model it using conditional random fields (CRFs) on top of hand-engineered features \cite{finkel2005incorporating} or bi-directional Long Short Term Memory Networks (LSTMs) \cite{lample2016neural, chiu2016named}. Recently, NER systems have been achieving state-of-the-art results by using word contextual embeddings, obtained with language models \cite{Peters2018, devlin2018bert, Akbik_flair}. 

Most EL systems discard mention detection, performing only entity disambiguation of previously detected mentions. Thus, in these cases the dependency between the two tasks is ignored. EL state-of-the-art methods often correspond to local methods which use as main features a candidate entity representation, a mention representation, and a representation of the mention's context \cite{sun2015modeling, yamada2016joint, YamadaSTT17, ganea2017deep}. Recently, there has also been an increasing interest in attempting to improve EL performance by leveraging knowledge base information \cite{radhakrishnan2018elden} or by allying local and global features, using information about the neighbouring mentions and respective entities \cite{le2018improving, Cao_el, yang2018collective}. However, these approaches involve knowing the surrounding mentions which can be impractical in a real case because we might not have information about the following sentences. It also adds extraneous complexity that might implicate a longer time to process.

Some works, as in this paper, perform end-to-end EL trying to leverage the relatedness of mention detection or NER and EL, and obtained promising results. \citet{Kolitsas_el} proposed a model that performs mention detection instead of NER, not identifying the type of the detected mentions, as in our approach.  \citet{sil2013re}, \citet{luo2015joint}, and \citet{nguyen2016j} introduced models that do joint learning of NER and EL using hand-engineered features. \cite{durrett2014joint} performed joint learning of entity typing, EL, and coreference using a structured CRF, also with hand-engineered features. 
In contrast, in our model we perform multi-task learning \cite{caruana1997multitask, evgeniou2004regularized}, using learned features.

\section{Model Description}
\begin{figure*}[ht!]
\begin{center}
\includegraphics[width=16cm]{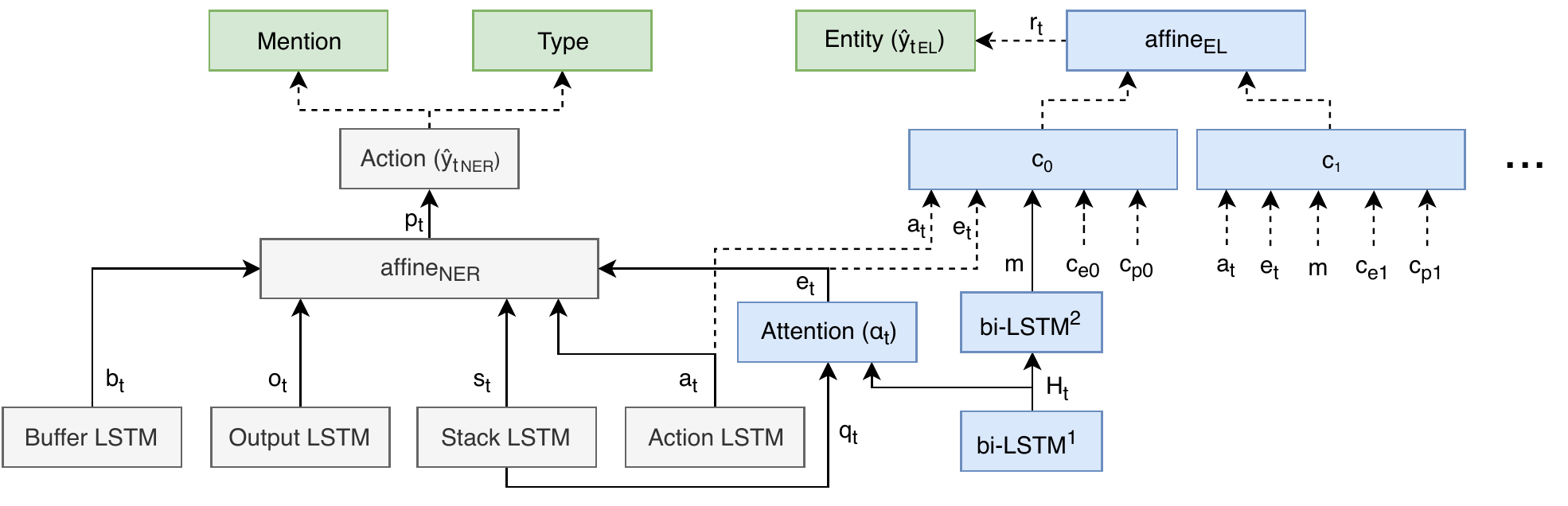}
\end{center}
\caption{Simplified diagram of our model. The dashed arrows only occur when the action is \textit{Reduce}. The blocks in blue correspond to our extensions to the Stack-LSTM and the green blocks correspond to the model's predictions. The grey blocks correspond to the stack-LSTM, the blue blocks to our extensions, and the green ones to the outputs.}
\label{diagram}
\end{figure*}

In this section firstly, we briefly explain the Stack-LSTM \citep{Dyer_stack_2015, lample2016neural}, model that inspired our system. Then we will give a detailed explanation of our modifications and of how we extended it to also perform EL, as showed in the diagram of Figure~\ref{diagram}. An example of how the model processes a sentence can be viewed in Table \ref{example_stack}.

\subsection{Stack-LSTM}
The Stack-LSTM corresponds to an action-based system which is composed by LSTMs augmented with a stack pointer.
In contrast to the most common approaches which detect the entity mentions for a whole sequence, with Stack-LSTMs the entity mentions are detected and classified on the fly. This is a fundamental property to our model, since we perform EL when a mention is detected.

This model is composed by four stacks: the \textit{Stack}, that contains the words that are being processed, the \textit{Output}, that is filled with the completed chunks, the \textit{Action} stack, which contains the previous actions performed during the processing of the current document, and the \textit{Buffer}, that contains the words to be processed. 

For NER, in the Stack-LSTM there are three possible types of actions: 

\begin{itemize}
    \item \textit{Shift}, that pops a word off the \textit{Buffer} and pushes it into the \textit{Stack}. It means that the last word of the \textit{Buffer} is part of a named entity.
    \item \textit{Out}, that pops a word off the \textit{buffer} and inserts it into the \textit{Output}. It means that the last word of the \textit{Buffer} is not part of a named entity.
    \item \textit{Reduce}, that pops all the words in the \textit{Stack} and pushes them into the  \textit{Output}. There is one action \textit{Reduce} for each possible type of named entities, e.g. \textit{Reduce-PER} and \textit{Reduce-LOC}.
\end{itemize}
Moreover, the actions that can be performed at each step are controlled: the action \textit{Out} can only occur if the stack is empty and the actions \textit{Reduce} are only available when the \textit{Stack} is not empty.

\subsection{Our model}

\paragraph{NER.} To better capture the context, we complement the Stack-LSTM with a representation $\bs{v}_t$ of the sentence being processed, for each action step $t$. For that the sentence $\bs{x}_1, \ldots, \bs{x}_{|w|}$ is passed through a bi-directional LSTM, being $\bs{h}_w^1$ the hidden state of its 1\textsuperscript{st} layer (bi-LSTM$^1$ in Figure~\ref{diagram}), that corresponds to the word with index $w$: 
\begin{align*}
    &\{\bs{h}_{1}^1,\dots,\bs{h}_{|w|}^1\} = \mathrm{BiLSTM}^1(\bs{x}_1,\dots,\bs{x}_{|w|}).\\
\end{align*}
We compute a representation of the words contained in the \textit{Stack}, $\bs{q}_t$, by doing the mean of the hidden states of the 1\textsuperscript{st} layer of the bi-LSTM that correspond to the words contained in the stack at action step $t$, set $\mathcal{S}_t$,:
\begin{align*}
    &\bs{q}_t = \dfrac{\sum_{k\in \mathcal{S}_t}\bs{h}_k^1}{|\mathcal{S}_t|}. \\
\end{align*}
This is used to compute the attention scores ${\bs{\alpha}_t}$:
\begin{align*}
    &{z_t}_w = \bs{u}^{\top} (\bs{W}_1 \bs{h}_{w}^1 + \bs{W}_2 \: \bs{q}_t)\\
    &\bs{\alpha}_t = \softmax(\bs{z}_t),\\
\end{align*}
where $\bs{W}_1$, $\bs{W}_2$, and $\bs{u}$ are trainable parameters. 
The representation $\bs{v}_t$ is then obtained by doing the weighted average of the bi-LSTM 1\textsuperscript{st} layer's hidden states:
\begin{align*}
    &\bs{v}_t = \sum_{w=1}^{|w|} \bs{h}_{w}^1 \: {\alpha_t}_w. \\
\end{align*}
To predict the action to be performed, we implement an affine transformation ($\text{affine}_\text{NER}$ in Figure~\ref{diagram}) whose input is the concatenation of the last hidden states of the \textit{Buffer} LSTM $\bs{b}_t$, \textit{Stack} LSTM $\bs{s}_t$, \textit{Action} LSTM $\bs{a}_t$, and \textit{Output} LSTM $\bs{o}_t$, as well as the sentence representation $\bs{v}_t$. 
\begin{align*}
    &\bs{d}_t=[\bs{{b}}_t;\; \bs{s}_{t};\; \bs{a}_{t};\; \bs{o}_{t};\; \bs{v}_t]\\
\end{align*}
Then, for each step $t$, we use these representations to compute the probability distribution $\bs{p}_t$ over the set of possible actions $\mathcal{A}$, and select the action $\widehat{\bs{y}}_{t_{\mathrm{NER}}}$ with the highest probability:

\begin{align*}
    &\bs{p}_t=\softmax(\affine(\bs{d}_t))\\
    &\widehat{\bs{y}}_{t_{\mathrm{NER}}} = \argmax_{a\: \in\: \mathcal{A}}\:(\bs{p}_t(a)).\\
\end{align*}
The NER loss function is the cross entropy, with the gold action for step $t$ being represented by the one-hot vector $\bs{y}_{t_{\mathrm{NER}}}$:

\begin{align*}
    \mathcal{L}_{NER} = -\sum_{t=1}^T \bs{y}_{t_{\mathrm{NER}}}^{\top} \log(\bs{p}_t).\\
\end{align*}
where $T$ is the total number of action steps for the current document.

\paragraph{EL.} When the action predicted is \textit{Reduce}, a mention is detected and classified. This mention is then disambiguated by selecting its respective entity knowledge base id. 
The disambiguation step is performed by ranking the mention's candidate entities.

The candidate entities ${c \in \mathcal{C}}$ for the present mention are represented by their entity embedding $\bs{c}_e$ and their prior probability $c_p$. The prior probabilities were previously computed based on the co-occurrences between mentions and candidates in Wikipedia.

To represent the mention detected the 2\textsuperscript{nd} layer of the sentence bi-LSTM (bi-LSTM$^2$ in Figure~\ref{diagram}), is used, being the representation $\bs{m}$ obtained by averaging the hidden states $\bs{h}_w^2$ that correspond to the words contained in the mention, set $\mathcal{M}$:

\begin{align*}
    &\{\bs{h}_{1}^2,\dots,\bs{h}_{|w|}^2\} = \mathrm{BiLSTM}^2(\bs{h}_1^1,\dots,\bs{h}_{|w|}^1)\\
    &\bs{m}=\dfrac{\sum_{w \in  \mathcal{M}} \; \bs{h}_w^2}{|\mathcal{M}|}. \\
\end{align*}
These features are concatenated with the representation of the sentence $\bs{v}_{t}$, and the last hidden state of the \textit{Action} stack-LSTM $\bs{a}_t$:
\begin{align*}
    \bs{c}_i = [\bs{c}_{ei};\; c_{pi};\; \bs{m};\; \bs{v}_{t};\; \bs{a}_t].\\
\end{align*}
We compute a score for each candidate with affine transformations ($\text{affine}_\text{EL}$ in Figure~\ref{diagram}) that have $\bs{c}$ as input, and select the candidate entity with the highest score,  $\widehat{\bs{y}}_{t_{\mathrm{EL}}}$:

\begin{align*}
    &\bs{l}_t = \affine(\tanh(\affine(\bs{c}_i, \dots, \bs{c}_n)))\\
    &\bs{r}_t = \softmax(\bs{l}_t)\\
    &\widehat{\bs{y}}_{t_{\mathrm{EL}}} = \argmax_{c\: \in\: \mathcal{C}}\:(\bs{r}_t(c)).\\
\end{align*}
The EL loss function is the cross entropy, with the gold entity for step $t$ being represented by the one-hot vector $\bs{y}_{t_{\mathrm{EL}}}$:

\begin{align*}
    \mathcal{L}_{EL} = -\sum_{t=1}^T \bs{y}_{t_{\mathrm{EL}}}^{\top} \log(\bs{r}_t)).\\
\end{align*}
where $T$ is the total number of mention that correspond to entities in the knowledge base.

Due to the fact that not every mention detected has a corresponding entity in the knowledge base, we first classify whether this mention contains an entry in the knowledge base using an affine transformation followed by a sigmoid. The affine's input is the stack LSTM last hidden state $\bs{s}_t$:
\begin{align*}
    d = \sigmoid(\affine(\bs{s}_t)).\\
\end{align*}
The NIL loss function, binary cross-entropy, is given by:

\begin{align*}
    \mathcal{L}_{NIL} =\! -(y_{NIL}\log(d)\!+\!(1-y_{NIL})\log(1-d)),\\
\end{align*}
where $y_{NIL}$ corresponds to the gold label, $1$ if mention should be linked and $0$ otherwise.

During training we perform teacher forcing, i.e. we use the gold labels for NER and the NIL classification, only performing EL when the gold action is \textit{Reduce} and the mention has a corresponding id in the knowledge base. The multi-task learning loss is then obtained by summing the individual losses:

\begin{align*}
    \mathcal{L} = \mathcal{L}_{NER}+\mathcal{L}_{EL}+\mathcal{L}_{NIL}.\\
\end{align*}

\section{Experiments}

\subsection{Datasets and metrics}
We trained and evaluated our model on the biggest NER-EL English dataset, the AIDA/CoNLL dataset \citep{Hoffart2011}. It is a collection of news wire articles from Reuters, composed by a training set of 18,448 linked mentions in 946 documents, a validation set of 4,791 mentions in 216 documents, and a test set of 4,485 mentions in 231 documents. In this dataset, the entity mentions are classified as person, location, organisation and miscellaneous. It also contains the knowledge base id of the respective entities in Wikipedia.

For the NER experiments we report the F1 score while for the EL we report the micro and macro F1 scores. The EL scores were obtained with the Gerbil benchmarking platform, which offers a reliable evaluation and comparison with the state-of-the-art models \cite{roder2017gerbil}. The results were obtained using strong matching settings, which requires exactly predicting
the gold mention boundaries and their corresponding entity.

\subsection{Training details and settings}

In our work, we used $100$ dimensional word embeddings pre-trained with structured skip-gram on the Gigaword corpus \cite{wang_sskipgr}. These were concatenated with $50$ dimensional character embeddings obtained using a bi-LSTM over the sentences. In addition, we use contextual embeddings obtained using a character bi-LSTM language model by \citet{Akbik_flair}.
The entity embeddings are $300$ dimensional and were trained by \citet{YamadaSTT17} on Wikipedia. To get the set of candidate entities to be ranked for each mention, we use a pre-built dictionary  \cite{pershina2015personalized}.

The LSTM used to extract the sentence and mention representations, $\bs{v}_t$ and $\bs{m}$ is composed by $2$ hidden layers with a size of $100$ and the ones used in the Stack-LSTM have $1$ hidden layer of size $100$. The feedforward layer used to determine the entity id has a size of $5000$. The affine layer used to predict whether the mention is NIL has a size of $100$. A dropout ratio of $0.3$ was used throughout the model. 

The model was trained using the ADAM optimiser \cite{kingma2014adam} with a decreasing learning rate of $0.001$ and a decay of $0.8$ and $0.999$ for the first and second momentum, respectively.

\subsection{Results}

\paragraph{Comparison with state of the art models.} We compared the results obtained using our joint learning approach with state-of-the-art NER models, in Table~\ref{ner_results}, and state-of-the-art end-to-end EL models, in Table~\ref{el_results}. In the comparisons, it can be observed that our model scores are competitive in both tasks. 

{
\renewcommand{\arraystretch}{1.2}
\begin{table}[h!]
\begin{centering}\small
\resizebox{\linewidth}{!}{%
\begin{tabular}{l@{\hskip 1cm}c}
\hline 
System &  Test F1\tabularnewline
\hline 
Flair \citep{Akbik_flair} & \textbf{93.09}\tabularnewline
BERT Large \citep{devlin2018bert} & 92.80 \tabularnewline
CVT + Multi \citep{clark_CVT} & 92.60 \tabularnewline
BERT Base \citep{devlin2018bert} & 92.40 \tabularnewline
BiLSTM-CRF+ELMo \citep{Peters2018} & 92.22  \tabularnewline
\hline
Our model & 92.43 \tabularnewline
\hline 
\end{tabular}}
\par\end{centering}
\centering{}\caption{NER results in CoNLL 2003 test set.}
\label{ner_results}
\end{table}
}

{
\renewcommand{\arraystretch}{1.2}
\begin{table}[h!]
\begin{centering}\small
\resizebox{\linewidth}{!}{%
\begin{tabular}{lcccc}
\hline 
System & \multicolumn{2}{c}{Validation F1} & \multicolumn{2}{c}{Test F1}\tabularnewline
& Macro & Micro & Macro & Micro \\
\hline 
\citet{Kolitsas_el} & \textbf{86.6} & \textbf{89.4} & \textbf{82.6} & \textbf{82.4} \tabularnewline
\citet{Cao_el} & 77.0 & 79.0 & 80.0 & 80.0 \tabularnewline
\citet{nguyen2016j} & - & - & - & 78.7 \tabularnewline
\hline
Our model &  82.8 & 85.2 & 81.2  & 81.9 \tabularnewline
\hline 
\end{tabular}}
\par\end{centering}
\centering{}
\caption{End-to-end EL results on validation and test sets in AIDA/CoNLL.}
\label{el_results}
\end{table}
}

\paragraph{Comparison with individual models.} To understand whether the multi-task learning approach is advantageous for NER and EL we compare the results obtained when using a multi-task learning objective with the results obtained by the same models when training with separate objectives. In the EL case, in order to perform a fair comparison, the mentions that are linked by the individual system correspond to the ones detected by the multi-task approach NER. 

These comparisons results can be found in Tables \ref{jlner} and \ref{jlel}, for NER and EL, respectively. They show that, as expected, doing joint learning improves both NER and EL results consistently. This indicates that by leveraging the relatedness of the tasks, we can achieve better models. 

{
\renewcommand{\arraystretch}{1.2}
\begin{table}[h!]
\begin{centering}\small
\begin{tabular}{l@{\hskip 0.8cm}p{21mm}p{13mm}}
\hline 
System &  Validation F1 & Test F1 \tabularnewline
\hline 
Only NER & 95.46 & 92.34  \tabularnewline

NER + EL & \textbf{95.72} & \textbf{92.52} \tabularnewline
\hline 
\end{tabular}
\par\end{centering}
\centering{}
\caption{Comparison of Named Entity Recognition multi-task results with single model results.}
\label{jlner}
\end{table}
}

{
\renewcommand{\arraystretch}{1.2}
\begin{table}[h!]
\begin{centering}\small
\begin{tabular}{lcccc}

\hline 
System &  \multicolumn{2}{c}{Validation F1} & \multicolumn{2}{c}{Test F1} \tabularnewline
& Macro & Micro & Macro & Micro \\
\hline 
Only EL & 81.3 & 83.5 &  79.9 & 80.2 \tabularnewline

NER + EL &  \textbf{82.6} & \textbf{85.2} & \textbf{81.1}  & \textbf{81.8} \tabularnewline
\hline 
\end{tabular}
\par\end{centering}
\centering{}
\caption{Comparison of Entity Linking results multi-task results with single model results.}
\label{jlel}
\end{table}
}

\paragraph{Ablation tests.} In order to comprehend which components had the greatest contribution to the obtained scores, we performed an ablation test for each task, which can be seen in Tables \ref{ablner} and \ref{ablel}, for NER and EL, respectively. These experiments show that the use of contextual embeddings (Flair) is responsible for a big boost in the NER performance and, consequently, in EL due to the better detection of mentions. We can also see that the addition of the sentence representation (sent rep $\bs{v}_t$) improves the NER performance slightly. Interestingly, the use of a mention representation (ment rep $\bs{m}$) for EL that is computed by the sentence LSTM, not only yields a big improvement on the EL task but also contributes to the improvement of the NER scores. The results also indicate that having a simple affine transformation selecting whether the mention should be linked, improves the EL results. 

{
\renewcommand{\arraystretch}{1.2}
\begin{table}[h]
\begin{centering}\small
\begin{tabular}{l@{\hskip 0.8cm}p{21mm}p{13mm}}
\hline 
System &  Validation F1 & Test F1 \tabularnewline
\hline 
Stack-LSTM                & 93.54  &  90.47 \tabularnewline
+ Flair                   & 95.40  &  92.16 \tabularnewline
+ sent rep                & 95.55  &  92.22 \tabularnewline
+ ment rep                & \textbf{95.72}  &  \textbf{92.52}\tabularnewline
+ NIL                     & 95.68  &  92.43\tabularnewline

\hline 
\end{tabular}
\par\end{centering}
\centering{}
\caption{Ablation test for Named Entity Recognition.}
\label{ablner}
\end{table}
}

{

\renewcommand{\arraystretch}{1.2}
\begin{table}[h!]
\begin{centering}\small
\begin{tabular}{lcccc}
\hline 
System &  \multicolumn{2}{c}{Validation F1} & \multicolumn{2}{c}{Test F1} \tabularnewline
& Macro & Micro & Macro & Micro \\
\hline 
Stack-LSTM                & 81.95 & 84.76 & 80.37 & 80.12 \tabularnewline
+ Flair                   & 82.59 & 85.75 & 80.86 & 81.05 \tabularnewline
+ sent rep                & 82.31 & 85.43 & 80.49 & 80.62 \tabularnewline
+ ment rep                & 82.64 & 85.17 & 81.07 & 81.76 \tabularnewline
+ NIL                     & \textbf{82.78} & \textbf{85.23} & \textbf{81.19} & \textbf{81.94}\tabularnewline
\hline 
\end{tabular}
\par\end{centering}
\centering{}
\caption{Ablation test for Entity Linking.}
\label{ablel}
\end{table}
}

\section{Conclusions and Future Work}
We proposed doing joint learning of NER and EL, in order to improve their performance. Results show that our model achieves results competitive with the state-of-the-art. Moreover, we verified that the models trained with the multi-task objective have a better performance than individual ones.
There is, however, further work that can be done to improve our system, such as training entity contextual embeddings and extending it to be cross-lingual.  

\section*{Acknowledgements}
This work was
supported by the European Research Council (ERC StG DeepSPIN 758969),
and by the Funda\c{c}\~ao para a Ci\^encia e Tecnologia
through contracts UID/EEA/50008/2019 and CMUPERI/TIC/0046/2014 (GoLocal). We thank Afonso Mendes, David Nogueira, Pedro Balage, Sebasti\~ao Miranda, Gon\c{c}alo Correia, Erick Fonseca, Vlad Niculae, Tsvetomila Mihaylova, Marcos Treviso, and the anonymous reviewers for helpful discussion and feedback.

\bibliography{acl2019}
\bibliographystyle{acl_natbib}

\end{document}